\documentclass[letterpaper, 10 pt, conference]{ieeeconf}  % Comment this line out if you need a4paper

\usepackage{graphicx}
\usepackage{multirow}
\usepackage{xcolor,colortbl} 
\usepackage[utf8]{inputenc}
\usepackage{pifont}
\usepackage[font=small,textfont=normal]{caption}
\usepackage{cite} 
\usepackage{makecell}
\usepackage{array}
\newcolumntype{M}[1]{>{\centering\arraybackslash}m{#1}}
\usepackage[colorlinks=true,citecolor=blue,linkcolor=blue,urlcolor=blue]{hyperref}

\IEEEoverridecommandlockouts                              % This command is only needed if 
\title{\LARGE \bf
FailSafe: Reasoning and Recovery from Failures in Vision-Language-Action Models
}

\author{
Zijun Lin$^{1,2}$ \quad
Jiafei Duan$^{3,4}$ \quad
Haoquan Fang$^{3,4}$ \\
Dieter Fox$^{3,4}$ \quad
Ranjay Krishna$^{3,4}$ \quad
Cheston Tan$^{2}$ \quad
Bihan Wen$^{1}$ \\
\\
$^{1}$Nanyang Technological University \quad
$^{2}$Centre for Frontier AI Research, A*STAR \\
$^{3}$Allen Institute for AI \quad
$^{4}$University of Washington \\
}

\begin{document}

\definecolor{lightorange}{RGB}{254, 225, 200}
\definecolor{darkgreen}{rgb}{0,0.5,0}

\maketitle
\thispagestyle{empty}
\pagestyle{empty}
% ``''

%%%%%%%%%%%%%%%%%%%%%%%%%%%%%%%%%%%%%%%%%%%%%%%%%%%%%%%%%%%%%%%%%%%%%%%%%%%%%%%%
\begin{abstract}
Recent advances in robotic manipulation have integrated low-level robotic control into Vision-Language Models (VLMs), extending them into Vision-Language-Action (VLA) models. Although state-of-the-art VLAs achieve strong performance in downstream robotic applications, supported by large-scale crowd-sourced robot training data, they still inevitably encounter failures during execution. Enabling robots to reason and recover from unpredictable and abrupt failures remains a critical challenge. Existing robotic manipulation datasets, collected in either simulation or the real world, primarily provide only ground-truth trajectories, leaving robots unable to recover once failures occur. Moreover, the few datasets that address failure detection typically offer only textual explanations, which are difficult to utilize directly in VLA models. To address this gap, we introduce FailSafe, a novel failure generation and recovery system that automatically produces diverse failure cases paired with executable recovery actions. FailSafe can be easily adapted to a wide range of manipulation tasks in simulators with motion planning support, enabling scalable creation of failure–action data. To demonstrate its effectiveness, we fine-tune LLaVa-OneVision-7B (LLaVa-OV-7B) to build FailSafe-VLM. Experimental results show that FailSafe-VLM successfully helps robotic arms detect and recover from potential failures, improving the performance of three state-of-the-art VLA models ($\pi_o$-FAST, OpenVLA, OpenVLA-OFT) by up to 22.6\% on average across several tasks in Maniskill. Furthermore, FailSafe-VLM could generalize across different spatial configurations, camera viewpoints, object and robotic embodiments. We plan to release the FailSafe code to the community. Project Page: https://jimntu.github.io/FailSafe/

% Project Page: \href{https://jimntu.github.io/FailSafe/}{\normalsize\textcolor{blue}{{https://jimntu.github.io/FailSafe/}}}

\end{abstract}

%%%%%%%%%%%%%%%%%%%%%%%%%%%%%%%%%%%%%%%%%%%%%%%%%%%%%%%%%%%%%%%%%%%%%%%%%%%%%%%%
\section{Introduction}

% action candidate (LLM zero-shot)
% textual explanation (visual summary, scene graph)
% only failure reasoning without recovery action

Vision-Language-Action (VLA) model has made remarkable progress recently in open-world robot manipulation task \cite{pi0_fast,openvla,openvla-oft,pi0,rdt,rt2}. Given the image observation and the language instruction, a VLA model can directly output executable robot actions to perform diverse tasks. The cornerstone of this progress lies in the increasingly large, high-quality robot datasets collected through community efforts \cite{bridgedata,khazatsky2024droid,openxembodiment, lin2025groundflow}. These datasets, either rolled out in simulation or teleoperated in real-world, typically consist of clean, ground-truth trajectories. However, relying solely on such correct data may not be sufficient, as robots inevitably make mistakes and may encounter situations not represented in the collected trajectories. Therefore, enabling robots to detect and recover from mistakes is crucial for achieving more robust and explainable downstream robotic applications.

Learning and recovering from failure is a fundamental aspect of human intelligence \cite{heyman2008children}, yet remains challenging to integrate into VLA models. Existing failure reflection methods for robotic manipulation, such as OLAF \cite{olaf} and YAY \cite{yell}, require an external observer to continuously monitor robot operation and intervene whenever a potential failure is detected. This reliance on human supervision is impractical for real-world scenarios where robots are expected to operate fully autonomously. 

\begin{figure}[t]
    \centering
    \includegraphics[width=0.9\linewidth]{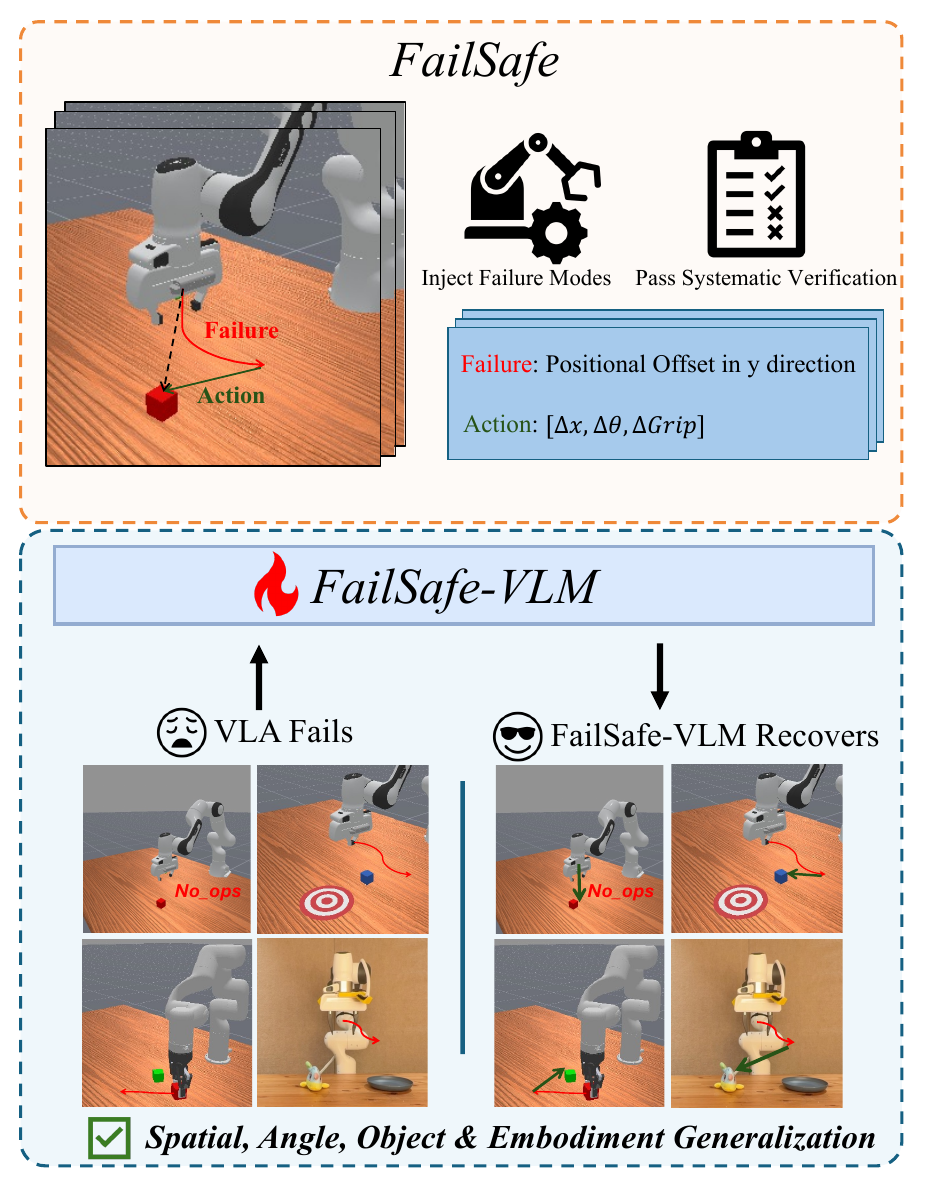}
    \caption{An illustration of the FailSafe pipeline generating failure scenarios and corresponding executable recovery actions (above). Leveraging these, FailSafe enables FailSafe-VLM (below) to detect and recover from robot failures, while generalizing across different spatial configurations, viewing angles, object and embodiments.}
    \label{fig:teaser}
\end{figure}

Meanwhile, recent pipelines focusing on robotic failures generation, such as AHA \cite{aha} and RoboFAC \cite{robofac}, are able to create large-scale failure trajectories dataset with detailed textual descriptions of task understanding and failure analysis. VLMs fine-tuned on these datasets have shown promising results in detecting potential failures automatically during task execution without human supervision. However, these pipelines fall short in failure correction, either overlooking this aspect entirely \cite{aha} or offering only textual feedback that robots cannot directly execute \cite{robofac, reflect}. For example, recovery instructions like ``the gripper should move left to align with the center of the cube'' are inherently ambiguous about magnitudes, scales and endpoints, thereby limiting their effectiveness in improving VLA control. This raises a key question: \textit{Is it possible to design an automatic pipeline that generates both high-level failure reasoning and low-level corrective actions at scale, in a way that directly benefits VLA models?}

To address this gap, we propose FailSafe, an automatic failure generation and recovery pipeline designed to adapt seamlessly across diverse tasks, simulators, and embodiments. As illustrated in Figure \ref{fig:teaser}, the dataset produced by FailSafe comprises two primary components: possible Failure scenarios and their corresponding executable recovery actions. FailSafe introduces diverse failure types including translation, rotation and no-ops failure with randomly sampled perturbation magnitudes. These failures are injected at arbitrary steps of ground-truth rollouts in simulation, intentionally causing the original task to fail. This design closely mimics how VLA-controlled robots can make mistakes unexpectedly during real deployment. 

Furthermore, unlike prior work that frames failure reasoning merely as textual explanations \cite{sharma2022correcting,aha,robofac}, FailSafe collects executable recovery actions that can be directly applied by robots in real time. To ensure accuracy and robustness, each recovery action is validated through a rigorous systematic verification, confirming that it effectively resolves the failure scenario and eventually leads to successful task completion. By incorporating multi-view images and task instructions, this pipeline produces a customized, accurate, and explainable FailSafe dataset for robotic manipulation that would be prohibitively difficult to obtain through real-world data collection.

Experimental results further highlight the effectiveness of FailSafe in VLA deployments. We fine-tune LLaVa-OV-7B \cite{llava-ov} to obtain FailSafe-VLM, which demonstrates strong failure reasoning capabilities that generalize across diverse spatial configurations, camera viewpoints, different objects and robotic embodiments. When serving as an external assistant to VLA models, FailSafe-VLM identifies and corrects potential failures in real time during robot execution, leading to an average performance improvement of up to 22.6\% on ManiSkill tasks \cite{maniskill3} compared to three VLA baselines without FailSafe-VLM. Furthermore, FailSafe-VLM significantly outperforms state-of-the-art VLMs, such as GPT-4o \cite{gpt4} and Gemini-2.5-flash \cite{gemini}, in detecting failures and generating accurate recovery actions on unseen failure trajectories.

Overall, we make the following contributions.

 \begin{itemize}
     \item We are the first to propose FailSafe, a scalable framework that can be adapted to build on top of simulators with motion planning to generate both failure reasoning explanations and accurate recovery actions that can be directly executed by robots.
    \item We show that FailSafe dataset enables existing VLMs to reason about failures and significantly improves the performance of three VLA models with a small inference overhead.
    \item We conduct extensive experiments to demonstrate that FailSafe-VLM could be generalized across diverse camera angles, object categories and robotic embodiments.
 \end{itemize}

% The intuition behind using end effector instead of joint rotation is that the corrective action could be directly applied to robots with any joint configuration.
\section{Related Work}

\subsection{Vision-Language-Action model}
Due to community efforts in crowd-sourcing robot learning data, VLM models with spatial reasoning capabilities trained on these datasets are now able to produce accurate, executable commands that can directly control robots, marking the emergence of VLA models \cite{bjorck2025gr00t,bu2025univla,li2025bridgevla}. Given image observations and task instructions, VLA models can generate low-level actions that directly control joint angles or the end-effector of the robots. Early designs of VLA models, such as OpenVLA \cite{openvla}, leverage  Prismatic-7B \cite{karamcheti2024prismatic} fine-tuned on the Open-X-Embodiment dataset \cite{openxembodiment} to predict robot actions. By using the language model's tokenizer, they map continuous robot actions into 256 of the least-used discrete tokens, framing action prediction as a ``vision-language'' task. While this approach demonstrates promising capabilities in robot manipulation, the discrete action token design conflicts with the inherently continuous nature of robot actions, limiting the effectiveness of the method especially when the tasks demand dexterity.

Several approaches have been proposed to address the limitations of discrete action tokens. OpenVLA-OFT \cite{openvla-oft} introduces action chunking and regression loss, replacing the original next-token prediction and cross-entropy loss of OpenVLA. Additionally, drawing inspiration from image generation, models such as $\pi_o$-FAST \cite{pi0_fast} and Diffusion-VLA \cite{wen2025diffusionvla} adopt an expert action head and integrate techniques like diffusion or flow matching with a VLM backbone, enabling them to generate continuous, high-frequency actions. More recently, VLA models with reasoning capabilities have emerged \cite{lee2025molmoact,zhou2025chatvla,huang2025thinkact}. Instead of directly outputting executable actions, these models generate reasoning signals, such as intermediate sub-goals, predicted trajectories, or other mid-level representations, to improve both the robustness and the explainability of VLA models.

However, existing robot learning datasets and crafted reasoning datasets mostly are built solely on clean, correct trajectories. While this allows robots to perform well in scenarios similar to the training distribution, real-world execution is imperfect, and failures are inevitable. Hence, there is a need to incorporate failure recovery information into current datasets so that VLAs can develop failure reasoning capabilities. To this end, we propose a novel pipeline that automatically generates failure data along with corresponding recovery actions, enabling VLAs to reason about and recover from failure scenarios.

\begin{figure*}[t]
    \centering
    \includegraphics[width=0.95\textwidth]{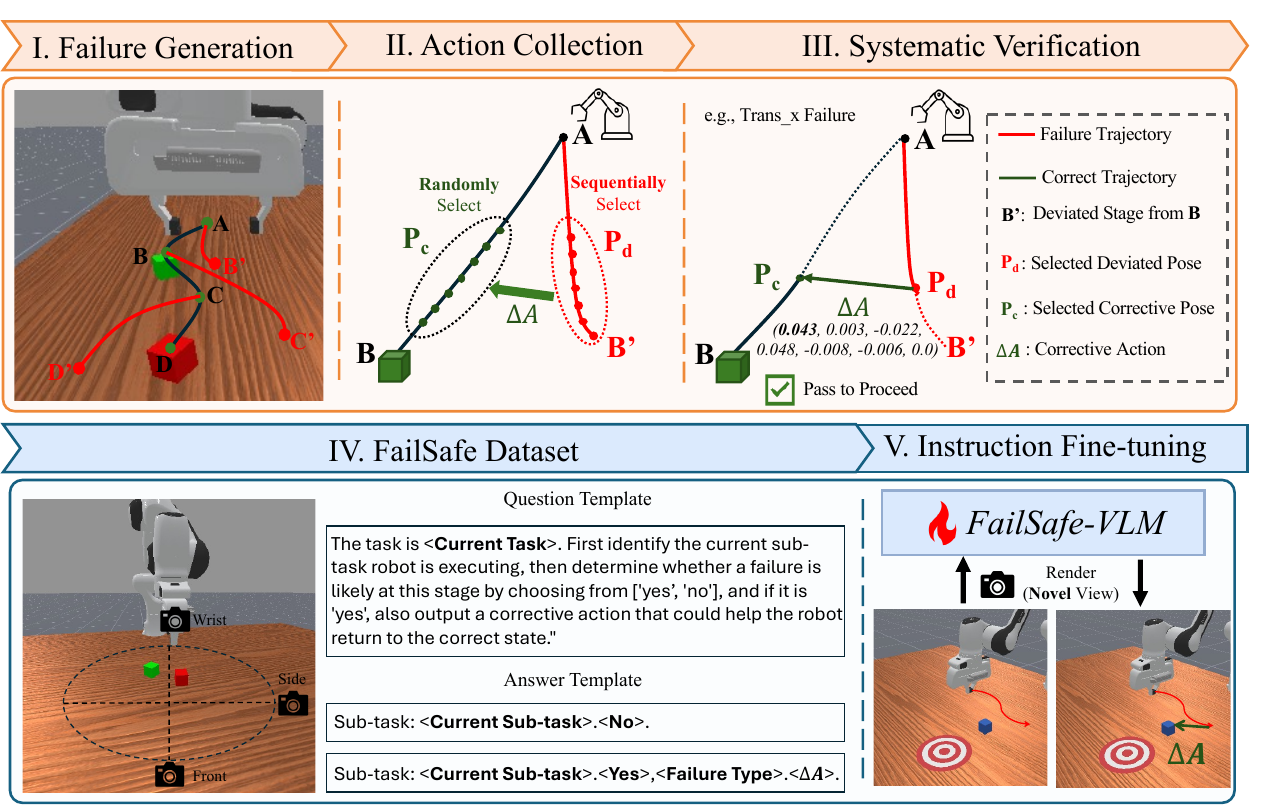}
      \caption{\textcolor{orange}{Top}: Overall pipeline of FailSafe, which includes the autonomous generation of failure trajectories (I) and collection of delta recovery action (II). Failure-Action data pairs are passed to the next step only after a systematic verification (III) ensures the effectiveness of recovery action. \textcolor{blue}{Bottom}: The FailSafe dataset (IV) is then used to fine-tune FailSafe-VLM, which is able to help robotic arms recover from failure cases (V).}
  \label{fig:FailSafe}
\end{figure*}

\subsection{Failure reasoning on robotic manipulation}

Early attempts to equip robot policies with failure reasoning relied on external human observers. For example, OLAF \cite{olaf} collected failure recovery synthesis data by prompting a large language model (LLM) with human verbal corrections and state observations, requiring the model to choose one corrective action from candidates. Similarly, YAY \cite{yell} used human intervention to update its high-level language policy. However, the failure data collected through such approaches are difficult to generate at scale and often lack accuracy.

To automate this process, REFLECT \cite{reflect} built the RoboFail dataset by prompting an LLM with hierarchical robot summaries for each sub-goal and letting the LLM determine whether a potential failure had occurred. Moreover, AHA \cite{aha} and RoboFAC \cite{robofac} systematically perturbed key poses in simulators to generate failure scenarios. While these methods represent an important step toward scalable failure reasoning data compared to earlier human-involved approaches, their failure correction still relies on natural language instructions \cite{dai2025racer,chen2024automating,zawalski2024robotic}. In these cases, corrections are appended to the original task prompt, forming a revised textual prompt. However, VLMs have limited capacity to follow complex natural language instructions, and such corrections cannot be directly applied to current VLA models, potentially restricting their flexibility in assisting robot control. 

Therefore, generating recovery actions that can be executed directly by robots is essential for producing tangible benefits in manipulation tasks. In this work, FailSafe advances the field by automatically generating scalable failure reasoning data along with corresponding 7-DoF end-effector recovery actions, directly enhancing VLA control.

\section{FailSafe}

We propose a novel pipeline, FailSafe, to systematically collect failure cases and corresponding recovery actions within ManiSkill \cite{maniskill3}. While developed in ManiSkill, this pipeline is designed to generalize across a broader range of simulators that support motion planning, where valid trajectories are automatically generated between designated start and goal poses. To ensure the reliability of the collected data, we incorporate a systematic verification mechanism that validates each failure–action pair before inclusion in the dataset. This step is crucial to guarantee both the diversity and effectiveness of the recovery actions. Details of the failure generation, action collection and systematic verification mechanism are provided in Section \ref{subsec:failure_gen}, Section \ref{subsec:action_collection} and Section \ref{subsec:sanity_check}, respectively.

FailSafe dataset is subsequently used to fine-tune LLaVa-OV-7B \cite{llava-ov}, yielding FailSafe-VLM, which is an expert vision–language model capable of assisting VLA models in reasoning and recovering from failures during robotic manipulation. The dataset format and the fine-tuning process for obtaining FailSafe-VLM are described in Section \ref{subsec:dataset} and Section \ref{subsec:tune}, respectively.

\subsection{Failure Generation}
\label{subsec:failure_gen}

To model failures that are frequently encountered in robotic manipulation tasks, we define three basic failure modes: \textbf{translation} failure, \textbf{rotation} failure, and \textbf{no-ops} failure. Translation failures correspond to perturbations along the Cartesian axes $(x,y,z)$, while rotation failures represent angular deviations in roll, pitch, or yaw. Furthermore, no-ops failure is defined as the robotic arm becoming stuck for a certain period without any movement. 

In ManiSkill, as in most simulators \cite{rlbench, maniskill3, geng2025roboverse, chen2025robotwin}, motion planning decomposes a task into multiple stages, with a motion planner sequentially moving the robotic arm through the poses defined for each stage to complete the task. We leverage this setting to dynamically inject failure modes during task execution. All possible failure modes, noise ranges, and the stages at which failures may be introduced are specified in a YAML configuration file. Specifically, the noise range is set to $\pm0.1$ for translation failures and $\pm1$ radian for rotation failures. FailSafe utilizes this YAML file together with a customized environment wrapper to wrap around Maniskill. In this way, the pre-defined poses at each stage can be randomly perturbed with varying magnitudes, and each failure trajectory is designed to contain one deviated stage (i.e., from $B$ to $B'$). Consequently, the roll-out motion of the robotic arm becomes $A \rightarrow B' \rightarrow C \rightarrow D$, as illustrated in Figure~\ref{fig:FailSafe}. If the task ultimately fails due to our introduced perturbation, FailSafe automatically records the image observations, failure trajectory, and failure type, before passing the case to the action collection phase.

Notably, the defined failure modes are simple yet fundamental. Many multi-step failures, such as an object slipping during transport, can be traced back to initial improper grasps caused by basic translation or rotation deviation. Hence, the FailSafe pipeline explicitly considers these delayed-failure cases, where the task fails several steps after the root error occurs. Moreover, our framework also handles cases where multiple failures occur simultaneously, as the corrective action $\Delta A$ provides meaningful adjustments across all seven dimensions, even though only one main failure type is specified. Further details regarding corrective action are provided in Section \ref{subsec:action_collection}. Overall, these failure modes collectively offer a concise and comprehensive representation of most motion-level failures in VLA control \cite{aha}.

\subsection{Action Collection}
\label{subsec:action_collection}

Unlike previous pipelines that treat failure reasoning as textual explanations, FailSafe steps further to collect recovery actions. However, obtaining valid actions is non-trivial, since naively using the perturbation value as a delta action could cause collisions between the gripper and the object. Moreover, rather than being restricted to a fixed stage or timestep, corrections are expected to be applicable at any point before the failure fully unfolds. Therefore, a customized pipeline for recovery action collection is introduced in FailSafe. 

As shown in Figure \ref{fig:FailSafe}, for each pair of correct and failure trajectories of a specific stage, FailSafe aims to collect multiple candidate corrective actions $\Delta A$ that can be directly executed by robots. Each step in the trajectory is represented by a 7-DoF pose. Because failures are difficult to detect in the early steps, the search for corrections of deviated pose $P_d$ starts from the 10th step of the failure trajectory and continues until the final step (red dotted ellipse). Each deviated pose $P_d$ in failure trajectory is mapped to a corresponding corrective pose $P_c$ in the correct trajectory. To prevent potential collisions between the gripper and the object, this mapping is restricted to a window spanning from 10 steps after the start to 3 steps before the end of the correct trajectory (green dotted ellipse). For the no-op failure case, we randomly sample $P_c$ from 3 to 10 steps after $P_d$ along the correct trajectory. 

To obtain multiple corrections for a trajectory pair, the deviated pose $P_d$ is sequentially traversed across all candidates and randomly matched with a corrective pose $P_c$. This process yields several $(P_d, P_c)$ pairs along with their corresponding corrective actions $\Delta A$, computed as the 7-DoF difference between the two poses. In this way, several candidate $\Delta A$ are generated for recovering the robotic arm from a failure trajectory.

We would like to emphasize that although a specific failure type is used to generate the failure trajectory, the resulting corrective action $\Delta A$ is not necessarily 1-sparse. As illustrated in Figure \ref{fig:FailSafe}, due to the random sampling of the corrective pose, $\Delta A$ provides meaningful adjustments across all seven dimensions. The purpose of defining a failure type is to identify the dominant source of error, rather than to restrict $\Delta A$ to correcting only a single dimension. This design is better suited to assist VLA models, where multiple failures often occur simultaneously.

\begin{table}[t]
\centering
\setlength{\arrayrulewidth}{0.8pt}
% \scriptsize
\begin{tabular}{l|c|c|c|c}
\hline
Task & No-ops & Trans\_x & Trans\_y & Trans\_z \\ \hline
Pick Cube  & 7,485  & 10,575 & 5,295  & 0    \\
Push Cube  & 12,057 & 2,394  & 13,947 & 2,385  \\
Stack Cube & 6,693  & 11,511 & 9,792  & 0     \\ \hline
\textbf{Total} & \textbf{26,235} & \textbf{24,480} & \textbf{29,034} & \textbf{2,385} \\ \hline \hline
Task  & Rot\_x & Rot\_y & Rot\_z & GT \\ \hline
Pick Cube    & 60    & 69    & 60   & 24,351 \\
Push Cube & 15,690 & 11,397 & 2,565 & 16,893 \\
Stack Cube & 12,057 & 6,270  & 738  & 14,717 \\ \hline
\textbf{Total} & \textbf{27,807} & \textbf{17,736} & \textbf{3,363} & \textbf{55,961} \\ \hline
\end{tabular}
\caption{Distribution of failure and Ground-Truth (GT) entries across tasks collected by the FailSafe pipeline, yielding a failure-to-success ratio of 2.3:1.}
\label{tab:dataset}
\end{table}

\subsection{Systematic Verification}
\label{subsec:sanity_check}
After getting the candidate corrective actions, they need to go through rigorous systematic verification to be included into the final FailSafe dataset. Importantly, the systematic verification is designed to ensure that the collected recovery actions $\Delta A$ are both robust in correcting failure cases and effective when applied at any timestep where failures could potentially occur in the immediate future.

In systematic verification, FailSafe replays the trajectory by incorporating the two poses ($P_d$ and $P_c$) collected earlier. Specifically, the motion planner first moves the robotic arm to the deviated pose $P_d$, then to the corrective pose $P_c$, and finally continues with the subsequent poses of the task (i.e., $A \rightarrow P_d \rightarrow P_c \rightarrow B\rightarrow C \rightarrow D$). If the originally failed task now succeeds after applying the corrective pose $P_c$, the corrective action $\Delta A$ along with the information such as image observation and failure types collected during the failure generation phase is added to the dataset. This process ensures both the reliability and diversity of the collected recovery actions, thereby enhancing FailSafe-VLM’s ability to help VLA models recover from failure scenarios.

\subsection{FailSafe Dataset}
\label{subsec:dataset}
FailSafe pipeline is applied to collect failure scenarios and recovery actions for three tasks in Maniskill as an example, namely pick cube, push cube and stack cube, resulting in 131k failure-action pairs across various types of failures for different stages. Furthermore, approximately 56k ground-truth trajectory with no failures are also included to better enable the VLM differentiate between failure and successful cases. The statistics of FailSafe dataset is shown in Table \ref{tab:dataset}.

Figure \ref{fig:FailSafe} (IV) shows the detailed format of the dataset. It starts with the question asking the model whether there is a potential failure given the image observation and task instruction, then the answer includes current sub-task and whether there is a failure or not. If the answer is yes, identified failure type and corresponding recovery action are expected to output subsequently. Additionally, each failure or success entry contains 10 consecutive image observations of the robot trajectory. To provide a holistic view of the environment, three camera perspectives are included: front, side, and hand. However, during implementation shown in Figure \ref{fig:vlm}, a novel camera view aligned with the VLA training angle is introduced to evaluate whether FailSafe-VLM can generalize to unseen viewpoints and effectively collaborate with the VLA model in real-world settings.

\subsection{Instruction Fine-tuning}
\label{subsec:tune}
We performed full instruction fine-tuning of LLaVA-OneVision (7B) \cite{llava-ov} on the FailSafe dataset, co-training with a RoboPoint VQA mixture \cite{yuan2024robopoint} to improve generalization. Training ran for one epoch on 32 H100 GPUs with DeepSpeed ZeRO 3. The model was initialized from a single-image checkpoint, using Qwen2-7B-Instruct \cite{team2024qwen2} as the language backbone and SigLIP as the vision tower, with a two-layer GELU MLP projector (2× hidden expansion) and features from the vision encoder’s penultimate layer. We jointly fine-tuned the vision tower, MLP adapter, and language model with a base learning rate of 1e-5 (2e-6 for the vision tower), cosine decay with 3\% warmup, zero weight decay, and bfloat16/TF32 enabled.

\section{Experiments}
In this section, we design a series of experiments to evaluate the failure reasoning and recovery capabilities of FailSafe-VLM in Framka Emika Panda robot arm. First, we integrate FailSafe-VLM with three state-of-the-art VLA models to evaluate whether it can help them recover from failures and improve overall performance. Second, we investigate the generalization capability of FailSafe-VLM across different objects and robotic embodiments. Furthermore, we further roll out test seeds in Maniskill to generate unseen failure and success trajectories with new spatial configurations, and compare FailSafe-VLM with three other VLMs to assess the accuracy of predicted failure types and recovery actions. Finally, a qualitative analysis is conducted to visualize the end-effector pose when the VLA model is integrated with FailSafe-VLM.

\begin{figure}[t]
    \centering
    \includegraphics[width=\linewidth]{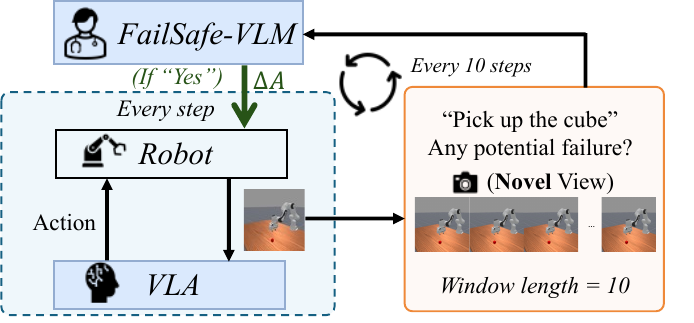}
    \caption{Illustration of how FailSafe-VLM collaborates with VLA models to perform failure reasoning and recovery. To simulate real-world settings, VLA models and FailSafe-VLM share the same camera view, which is used during VLA training but \textbf{novel} to FailSafe-VLM.}
    \label{fig:vlm}
\end{figure}

\subsection{Performance on VLA}
\label{subsec:vla_exp}
As shown in Figure \ref{fig:vlm}, FailSafe-VLM is designed to work side by side with VLA models. Every 10 steps, FailSafe-VLM takes over the base VLA model’s control of the robotic arm to detect potential failures and outputs actions that the robot can execute directly if a failure is identified. After that, control is returned to the base VLA model until the next 10 steps complete. 

To evaluate the effectiveness of the FailSafe pipeline in improving failure reasoning, we compare performance on three ManiSkill tasks for which FailSafe collects data on. Three commonly used autoregressive VLA models, $\pi_o$-FAST \cite{pi0_fast},  OpenVLA \cite{openvla}, and OpenVLA-OFT \cite{openvla-oft}, are also fine-tuned on 1,000 ground-truth trajectories for each task. All models are evaluated on test seeds, meaning the spatial configuration differs from the training environment. Moreover, the camera view used in the experiments is novel for FailSafe-VLM, while it matches the training view for the three baseline VLAs. The intuition behind this setting is to evaluate whether our pipeline can still accurately identify failure scenarios and assist VLA models when the observation viewpoint was not included in FailSafe pipeline. In real-world scenarios, an additional camera dedicated to FailSafe-VLM may not be available, making it reasonable to share the same camera angle with the base VLA models. Notably, our method directly outputs executable corrective actions for VLA control. In contrast, methods such as AHA \cite{aha} and RoboFAC \cite{robofac} produce only natural language instructions, which cannot be directly applied in this setting.

The experimental results in Table \ref{tab:exp} show that all three VLA models yield average performance improvements with the inclusion of FailSafe-VLM. In particular, OpenVLA improves by more than 20\%, with an average gain of 22.6\% across the three tasks. For OpenVLA-OFT and $\pi_o$-FAST, whose baselines are already relatively strong, failure reasoning still pushes the boundary, increasing success rates by 8.0\% and 4.0\%, respectively.

For seeds recovered by FailSafe-VLM, we observe helpful nudges when the robotic arm becomes stuck or is about to fail. We would like to highlight that FailSafe-VLM should be viewed as an assistant for failure reasoning and recovery, rather than as a replacement for VLA models. Overall, these results demonstrate that the FailSafe pipeline can automatically generate failure-recovery data at scale, adapt across multiple tasks and VLA models, and boost both the accuracy and robustness of robotic manipulation.
\begin{table}[t]
\centering
\setlength{\arrayrulewidth}{0.8pt}
\scriptsize

\begin{tabular}{c|c|ccc|c}
\hline
\makecell{VLA \\ models} & 
\makecell{FailSafe- \\VLM} & 
\makecell{Pick \\  Cube} & 
\makecell{Push \\  Cube} & 
\makecell{Stack \\ Cube} & 
\makecell{Average} \\
\hline
\multirow{3}{*}{$\pi_o$-FAST \cite{pi0_fast}}& \ding{55} & 88.0\% & 52.0\% & 96.0\%  & 78.7\%\\
& \ding{51} & 88.0\% & 64.0\% & 96.0\% & 82.7\%\\
& $\Delta$ & +0.0\% & \textcolor{darkgreen}{+12.0\%} & +0.0\% & \textcolor{darkgreen}{+4.0\%}\\
\hline
\multirow{3}{*}{OpenVLA \cite{openvla}} & \ding{55} & 28.0\% & 4.0\% & 12.0\% & 14.7\%\\
& \ding{51}& 48.0\% & 24.0\% & 40.0\% & 37.3\%\\
& $\Delta$ & \textcolor{darkgreen}{+20.0\%} & \textcolor{darkgreen}{+20.0\%} & \textcolor{darkgreen}{+28.0\%} & \textcolor{darkgreen}{+22.6\%}\\
\hline
\multirow{3}{*}{OpenVLA-OFT \cite{openvla-oft}}&\ding{55} & 84.0\% & 88.0\% & 100.0\% & 90.7\%\\
& \ding{51} & 96.0\% & 100.0\% & 100.0\% & 98.7\%\\
& $\Delta$ & \textcolor{darkgreen}{+12.0\%} & \textcolor{darkgreen}{+12.0\%} & +0.0\% & \textcolor{darkgreen}{+8.0\%}\\
\hline
\end{tabular}
\caption{Success rates of VLA models with and without FailSafe-VLM on three ManiSkill tasks using the Franka Emika Panda arm. Improvements are reported under the setting where the experimental camera view matches the VLA \textbf{training view} but is novel to FailSafe-VLM.}
\label{tab:exp}
\end{table}

\subsection{Generalization Capability}

To evaluate the generalization capability of FailSafe-VLM, we modify the objects used in three manipulation tasks. Specifically, Sphere and Charger are selected as novel object categories that FailSafe-VLM has never encountered during training. As shown in Table \ref{tab:object}, FailSafe-VLM successfully assists the VLA model in correcting failures even with these unseen objects, achieving an average improvement of 17.4\% across the three new tasks. This generalization ability can be attributed to the fundamental similarity of failure scenarios, as failure modes across different manipulation tasks often exhibit common patterns. The FailSafe pipeline enables the model to internalize these underlying failure principles, thereby empowering it to reason and recover from failures involving unseen objects in similar task settings.

We further evaluate whether FailSafe-VLM can assist VLA models in an embodiment unseen during training. Similarly, we also collect 1,000 trajectories per task on xArm 6 to fine-tune OpenVLA-OFT, while reusing the same FailSafe-VLM checkpoint trained exclusively on Franka Emika Panda robots. Although FailSafe-VLM is never trained on xArm 6 data, the failure scenarios and trajectories generated by FailSafe are independent of the specific robotic embodiment. Consequently, as shown in Table \ref{tab:embodiment}, FailSafe-VLM exhibits cross-embodiment generalization by boosting the performance of the stack cube task on xArm 6 from 56\% to 76\% without degrading performance on other tasks.

\subsection{Comparisons on other VLMs}
\label{subsec:roll-out}

To demonstrate that the failure reasoning capability stems from the carefully designed FailSafe pipeline rather than the inherent capabilities of the baseline VLMs, we further compare FailSafe-VLM with other state-of-the-art VLM models under a controlled setting.

Specifically, 20 seeds are held out as test seeds to generate both failure and success cases with spatial configurations different from the training data, resulting in 1,712 test entries. Three evaluation metrics are applied to comprehensively assess the failure reasoning ability of the VLMs. First, \textbf{binary success} is a two-class classification metric that measures whether the model can distinguish failure cases from success cases. Second, \textbf{accuracy} evaluates whether the model can correctly identify the specific failure types. An entry is counted as correct only if the model outputs the same failure type and axis as the ground truth. Finally, \textbf{cosine similarity} is computed between the ground-truth recovery action and the predicted recovery action to assess how well the predicted action aligns with the ground truth. To ensure a fair comparison, the other three VLM models are prompted with template with detailed instruction, a ground-truth example and the range of possible delta actions.

\begin{table}[t]
\centering
\setlength{\arrayrulewidth}{0.8pt}
\scriptsize
\begin{tabular}{c|c|ccc|c}
\hline
\makecell{VLA \\ model} & 
\makecell{FailSafe- \\VLM} & 
\makecell{Pick \\  \textbf{Sphere}} & 
\makecell{Place \\ \textbf{Sphere}} & 
\makecell{Pick \\ \textbf{Charger}} & 
\makecell{Average} \\
\hline
\multirow{3}{*}{OpenVLA-OFT \cite{openvla-oft}} &
\ding{55} & 44.0\% & 36.0\% & 80.0\% & 53.3\% \\
& \ding{51} & 68.0\% & 52.0\% & 92.0\% & 70.7\%\\
& $\Delta$ &  \textcolor{darkgreen}{+24.0\%}&  \textcolor{darkgreen}{+16.0\%} &  \textcolor{darkgreen}{+12.0\%} & \textcolor{darkgreen}{+17.4\%}\\
\hline
\end{tabular}
\caption{Success rates of OpenVLA-OFT with and without FailSafe-VLM for novel \textbf{object categories}.}
\label{tab:object}
\end{table}

\begin{table}[t]
\centering
\setlength{\arrayrulewidth}{0.8pt}
\scriptsize
\begin{tabular}{c|c|ccc|c}
\hline
\makecell{VLA \\ model} & 
\makecell{FailSafe- \\VLM} & 
\makecell{Pick \\  Cube} & 
\makecell{Push \\  Cube} & 
\makecell{Stack \\ Cube} & 
\makecell{Average} \\
\hline
\multirow{3}{*}{\makecell[c]{OpenVLA-OFT \cite{openvla-oft}\\\textbf{(xArm 6)}}} &
\ding{55} & 100.0\% & 100.0\% & 56.0\% & 85.3\% \\
& \ding{51} & 100.0\% & 100.0\% & 76.0\% & 92.0\%\\
& $\Delta$ & +0.0\% &  +0.0\%  &  \textcolor{darkgreen}{+20.0\%} & \textcolor{darkgreen}{+6.7\%}\\
\hline
\end{tabular}
\caption{Success rates of OpenVLA-OFT with and without FailSafe-VLM for novel \textbf{robotic embodiment}.}
\label{tab:embodiment}
\end{table}

\begin{table}[t]
\centering
\setlength{\arrayrulewidth}{0.8pt}
\scriptsize
\begin{tabular}{c|ccc}
\hline
VLM Models & Binary Success$\uparrow$ & Accuracy$\uparrow$& Cosine Similarity$\uparrow$\\
\hline
Qwen2.5-VL \cite{qwen-vl} & 0.2401 & 0.2401 & 0.0000\\
Gemini-2.5-flash \cite{gemini} & 0.6229 & 0.1412 & -0.0121\\
GPT-4o \cite{gpt4}& 0.7007 & 0.1960  & 0.0117\\
\textbf{FailSafe-VLM} & \textbf{0.9094} & \textbf{0.8368} & \textbf{0.6522}\\
\hline
\end{tabular}
\caption{Comparison of FailSafe-VLM with other state-of-the-art VLM models on failure reasoning and recovery, evaluated on roll-out test seeds in ManiSkill.}
\label{tab:baseline}
\end{table}

The results in Table \ref{tab:baseline} show that FailSafe-VLM significantly outperforms state-of-the-art VLM models across all three evaluation metrics. For example, Qwen2.5-VL consistently outputs ``no failure'' and an all-zero recovery action regardless of whether the trajectory actually fails, making it ineffective for assisting robots in recovery. Gemini-2.5-flash and GPT-4o perform reasonably well in detecting whether a failure has occurred (binary success), but both achieve less than 20\% accuracy when reasoning about the specific failure type. Their performance on recovery action prediction is also poor, with cosine similarity scores close to zero. 

In contrast, the FailSafe pipeline enables VLMs to achieve the highest binary success rate of 0.9094 and the best performance in both failure type reasoning and recovery action prediction. Specifically, FailSafe-VLM achieves over four times the accuracy of Gemini-2.5-flash and GPT-4o, and its cosine similarity exceeds 0.6. Importantly, strong failure reasoning in a VLM does not require near-perfect cosine similarity with the ground-truth action, since multiple corrective $\Delta A$ can enable recovery. Results in Section \ref{subsec:vla_exp} indicate that a cosine similarity of approximately $65\%$ already yields meaningful improvements. This demonstrates that FailSafe-VLM can detect failures under novel spatial configurations and generate effective recovery actions to correct failed trajectories.

\subsection{Inference Efficiency}
As shown in Table \ref{tab:speed}, the performance is significantly improved by up to 22.6\%, while incurring only marginal inference overheads of +3.8s to +9.1s. Most of the additional delay in the current system stems from simulator replanning after receiving corrective actions from FailSafe-VLM. The overall efficiency could be further improved with faster simulation or by adopting real-time action chunking techniques \cite{black2025real}. We leave these optimizations for future work.

\begin{table}[t]
\centering
\setlength{\arrayrulewidth}{0.8pt}
\scriptsize

\begin{tabular}{c|c|c|c}
\hline
\makecell{VLA \\ models} & 
\makecell{FailSafe- \\VLM} & 
\makecell{Performance} &
\makecell{Speed}\\
\hline
\multirow{3}{*}{$\pi_o$-FAST \cite{pi0_fast}}& \ding{55} &  78.7\% & 43.3s\\
& \ding{51} & 82.7\% & 47.2s\\
& $\Delta$ & \textcolor{darkgreen}{+4.0\%} & \textcolor{gray}{+3.9s}\\
\hline
\multirow{3}{*}{OpenVLA \cite{openvla}} & \ding{55} & 14.7\% & 112.1s\\
& \ding{51}&  37.3\% & 121.2s\\
& $\Delta$ &  \textcolor{darkgreen}{+22.6\%} & \textcolor{gray}{+9.1s}\\
\hline
\multirow{3}{*}{OpenVLA-OFT \cite{openvla-oft}}&\ding{55} & 90.7\% & 28.8s\\
& \ding{51} & 98.7\% & 32.6s\\
& $\Delta$ & \textcolor{darkgreen}{+8.0\%} & \textcolor{gray}{+3.8s}\\
\hline
\end{tabular}
\caption{Average performance and inference speed of VLA models across 75 runs with and without FailSafe-VLM.}
\label{tab:speed}
\end{table}

\subsection{Qualitative Analysis}

\begin{figure}[t]
    \centering
    \includegraphics[width=\linewidth]{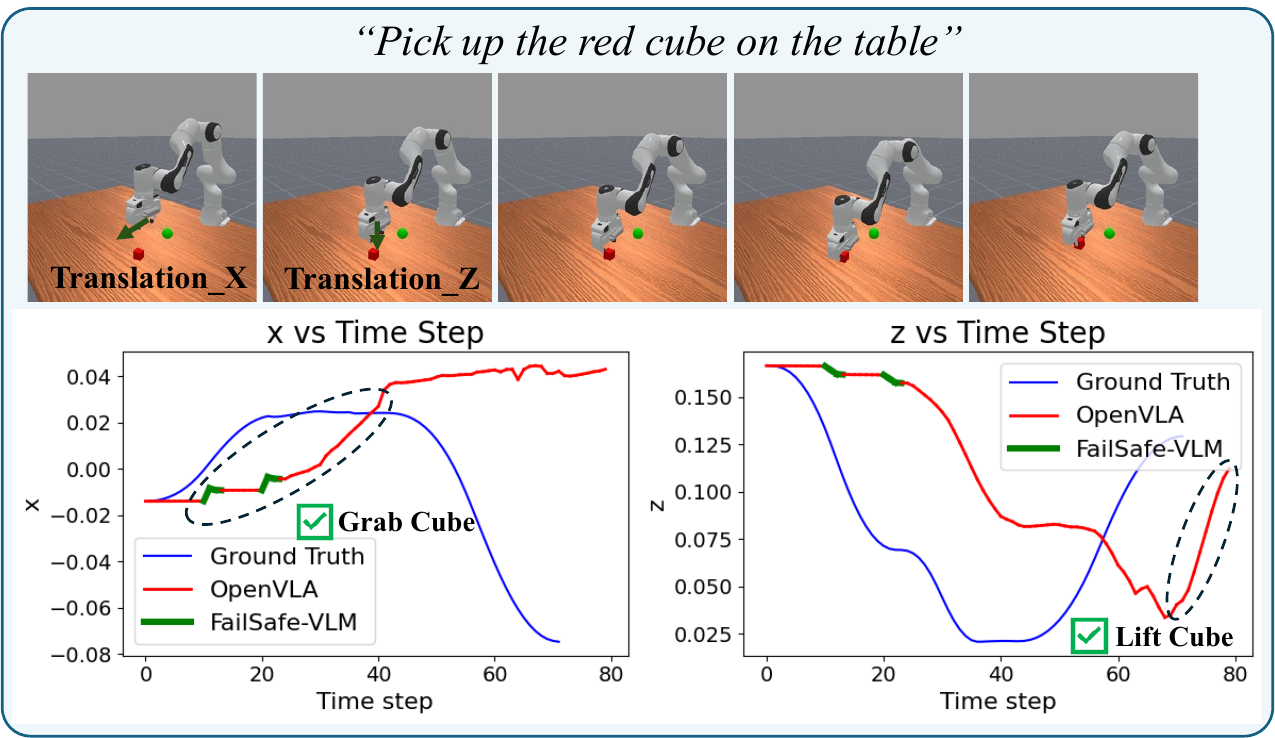}
    \caption{Examples of how FailSafe-VLM helps VLA models recover from failure scenarios, showing the x- and z-axis trajectories of the end effector over time (zoomed-in for clearer view).}
    \label{fig:vis}
\end{figure}

To better visualize the effect of FailSafe-VLM in failure recovery for robotic manipulation during VLA control, Figure \ref{fig:vis} shows how the x-axis and z-axis of the end effector change when FailSafe-VLM works together with OpenVLA. The corrections of the end effector pose made by FailSafe-VLM are highlighted in green. At the beginning, the robotic arm is nearly frozen. This failure mode would persist if the failure reasoning capability is not introduced, as the clean trajectories that VLA models are trained on do not include such scenarios, making self-recovery difficult. FailSafe-VLM, however, detects the potential failure and effectively nudges the arm closer to the ground-truth trajectory (green segments in Figure \ref{fig:vis}). The deviation of the x-axis from the ground-truth pose in the later part of the task is not critical, since once the arm reaches the cube’s x position (around 0.02 in the environment), further changes do not affect the ability to lift the cube. Finally, after guiding the arm back to the correct pose, OpenVLA resumes control and successfully completes the task, which could otherwise fail without the help of FailSafe-VLM. \textbf{Please refer to the supplementary video for robot demo.}

\section{Conclusions \& Discussions}
In this paper, we propose FailSafe, a novel pipeline automatically generates failure reasoning data with executable recovery actions. FailSafe can be readily extended to diverse manipulation tasks across current simulators that support motion planning, facilitating the scalable generation of failure datasets. Our experiments show that FailSafe-VLM exhibits strong failure reasoning capability and outperform other state-of-the-art VLMs. We further integrate FailSafe-VLM into VLAs to show its capability to recover from failures and boost overall performance in robotic manipulation tasks. FailSafe-VLM could also generalize cross spatial setups, camera views, object categories and robot embodiments. 

However, our method still has several limitations. The current FailSafe pipeline primarily focuses on motion-level recovery and does not yet support the correction of object-level errors. In addition, although integrating VLA with FailSafe-VLM yields performance gains, their synergy can be further improved in efficiency and flexibility through techniques such as real-time action chunking \cite{black2025real}. We leave these improvements for future work. In conclusion, we hope the FailSafe pipeline serves as an early attempt to enable stronger failure reasoning and autonomy in robot control systems, paving the way for more robust and explainable embodied AI applications.

\section*{Acknowledgements}

Zijun Lin is supported by the A*STAR Graduate Scholarship (Computing). This research is supported in part by A*STAR SERC CRF funding to C.T., and in part by A*STAR IAF-ICP Programme I2501E0041. The work was done at Rapid-Rich Object Search (ROSE) Lab, School of Electrical \& Electronic Engineering, Nanyang Technological University.

\bibliographystyle{IEEEtran}
%\bibliography{main}

\addtolength{\textheight}{-12cm}   % This command serves to balance the column lengths
                                  
\end{document}